\documentclass{article} 
\pdfoutput=1
\usepackage{iclr2021_conference,times,amsmath}
\usepackage{graphicx}
\usepackage{bm}

\usepackage{amsmath,amsfonts,bm}

\def\eqref#1{equation~\ref{#1}}

\def\1{\bm{1}}

\DeclareMathAlphabet{\mathsfit}{\encodingdefault}{\sfdefault}{m}{sl}
\SetMathAlphabet{\mathsfit}{bold}{\encodingdefault}{\sfdefault}{bx}{n}

\usepackage{hyperref}
\usepackage{url}

\title{Growing Neural Network with Shared \\Parameter}

\author{Ruilin Tong 
}

\iclrfinalcopy 
\begin{document}

	\maketitle
	
	\begin{abstract}
		We propose a general method for growing neural network with shared parameter by matching trained network to new input. By leveraging Hoeffding’s inequality, we provide a theoretical base for improving performance by adding subnetwork to existing network. With the theoretical base of adding new subnetwork, we implement a matching method to apply trained subnetwork of existing network to new input. Our method has shown the ability to improve performance with higher parameter efficiency. It can also be applied to trans-task case and realize transfer learning by changing the combination of subnetworks without training on new task.
	\end{abstract}
	
	\section{Introduction}
	
	Deep neural networks have made great progress and shown impressive empirical results recently. But most neural networks are trained in a fixed structure which limits the expandability of neural networks. To make neural networks expandable, there is a set of methods to optimize neural network structure dynamically. It has been shown that accurate and efficient neural network can be gradually grown from a relatively small network. (\cite{liu2019splitting}; \cite{wang2019energy}; \cite{wu2021firefly})
	
	\begin{figure}[h]
		\begin{center}
			\includegraphics[width=0.9\linewidth]{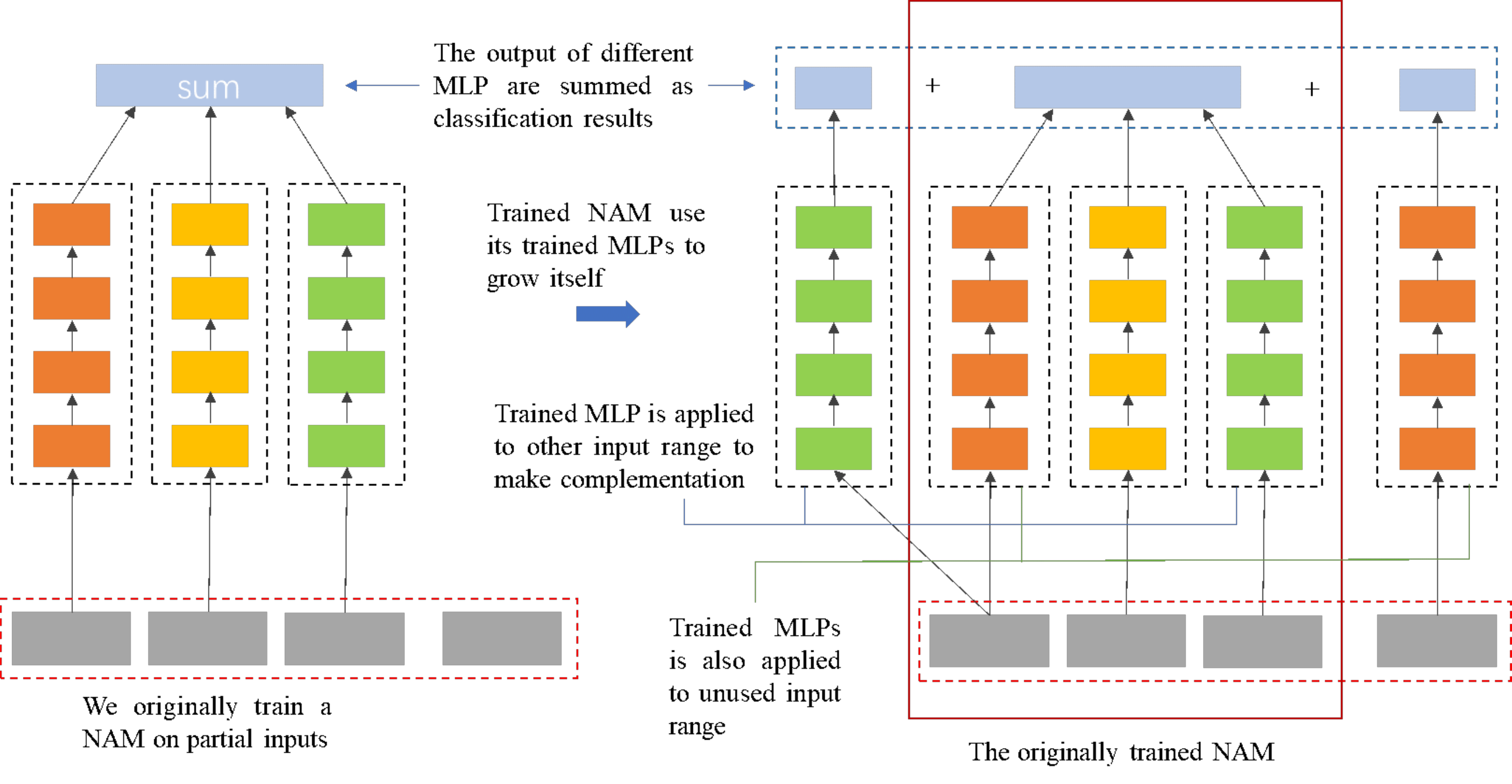}
		\end{center}
		\caption{General process of our growing method. We grow network by applying trained MLP to other input to improve performance. Each block represents a fully-connected layer, and sequence of blocks represents an MLP. MLPs in the same color is the same MLP.}
	\end{figure}
	
	The existing network growing methods grow networks by the indication of loss and grow neural network by adding new parameters. In this work, we propose a new method to grow neural networks with shared parameters based on statistic theory. The key problem of growing networks is that: under what condition, the added parameters is contributive to final results. We solve this problem by combing Hoeffding’s inequality and loss descendent in the case of summation ensemble, and this qualification method is expandable to multiple types of neural network. In the implementation, we apply the proposed qualification method to neural additive model (NAM) (\cite{agarwal2020neural}) which is the additive combination of multiple independent multi-layer perceptron (MLP), and grow neural network by matching trained MLP to new inputs. The general process is shown in Figure 1. Our growing method can also be applied in trans-task condition by matching trained MLP to the inputs of a new task as shown in Figure 2.
	
	\begin{figure}[h]
		\begin{center}
			\includegraphics[width=0.9\linewidth]{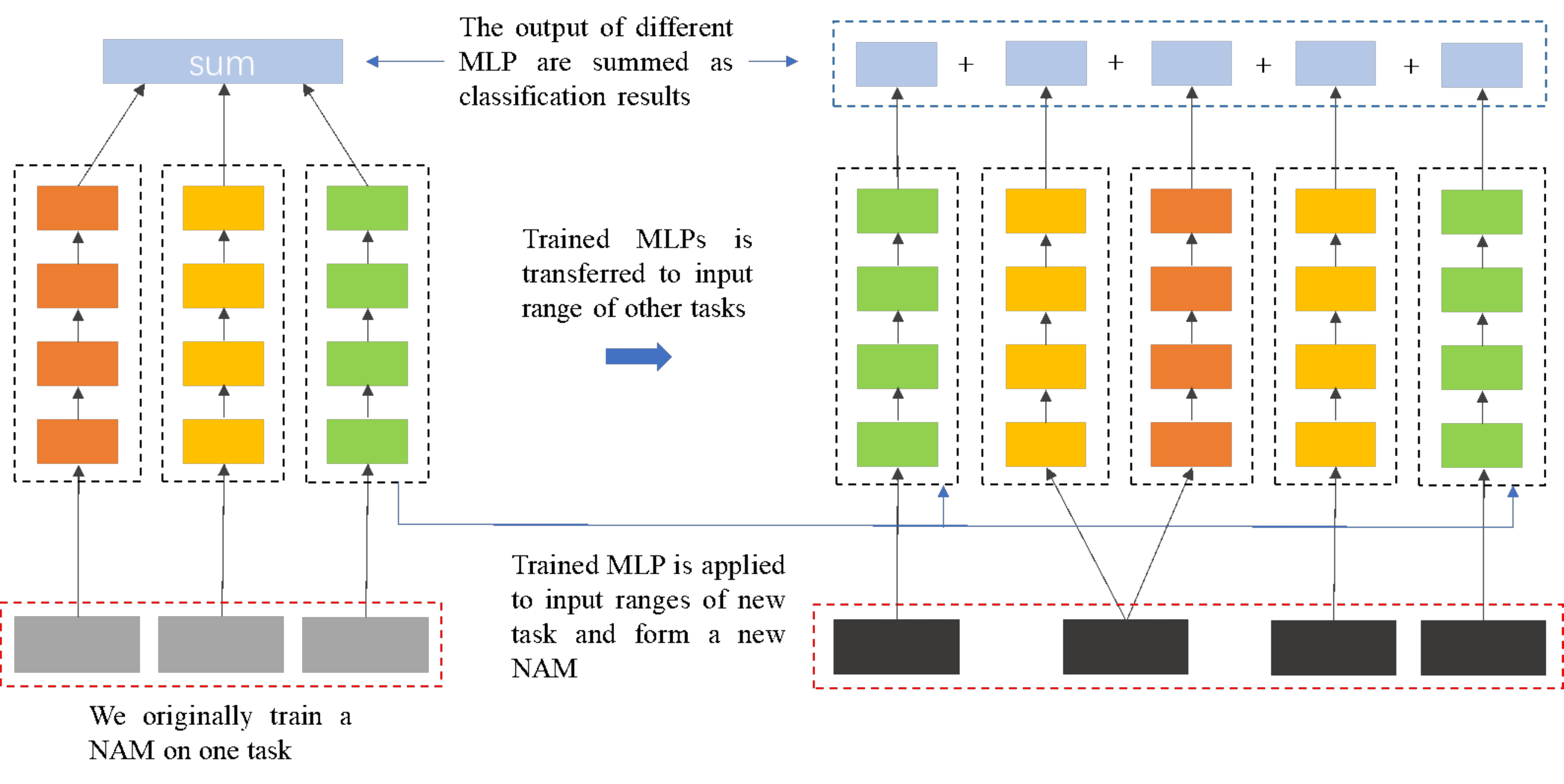}
		\end{center}
		\caption{Process of trans-task growing. We grow network by applying trained MLP to input of other tasks.}
	\end{figure}
	
	Our growing method is a general method and can be applied in additive ensemble networks on classification tasks. Theoretical base of MLP qualification guarantees a steady improvement on performance and matching trained MLP makes higher parameter efficiency. Our growing method can also be used under trans-task condition without training and without the requirement of shared knowledge between tasks. Experiment results have shown the steady increment of performance, higher parameter efficiency in growing existing network. We also show the ability for trans-task learning, which means that function of one neural network can be altered by perturbating parameters.
	
	\section{Methodology}
	
	\subsection{Subnetwork qualification}
	For a typical neural network on a classification task, the classification layer which is normally a fully-connected layer as shown in (1),\\
	\begin{equation}
	y_i=\sum_{j=1}^mw_{ij}f(x)_j+b_j\tag{1},
	\end{equation}
	where $\displaystyle f(x)_j$ is the output of former layers and $\displaystyle f$ could be any form of neural networks, $\displaystyle w_{ij}$ is the $\displaystyle ij$-th element of weight, $\displaystyle b_i$ is the $\displaystyle i$-th element of bias. In the rest part of this work, we refer $\displaystyle w_{ij}f(x)_j$ as the class-output of the $\displaystyle i$-th class, $\displaystyle y_i$ is the sum of the $\displaystyle i$-th class-output.
	
	From (1), we consider that class-output for each class is a linear combination of different values. Under this condition, we grow neural network by adding new subnetwork which add new value $\displaystyle w_{ij}f(x)_j$ to this linear combination. In this work, to make this problem easier, each new subnetwork only focuses on one class and the class-output of other classes is set to 0.
	
	The class which is focused is referred as target class and samples of target class is referred as target samples, other classes is referred as non-target class and samples of non-target class is referred as non-target samples.
	
	We evaluate the grown network on a dataset which is referred as selection set. The $\displaystyle j$-th sample of class-$\displaystyle i$ is represented as $\displaystyle sp_{ij}$, class-output of $\displaystyle k$-th subnetwork on sample $\displaystyle sp_{ij}$ is represented as $\displaystyle y_{ij}^k$, which is equal to $\displaystyle w_{ij}f(x)_j$ in (1). Number of classes is represented as $\displaystyle N_c$, number of subnetworks is represented as $\displaystyle N_n$.
	
	The evaluation considers two aspects, from loss and from Hoeffding’s inequality. Conditions for qualifying added subnetworks is summarized at the end of this section.
	
	\subsubsection{Hoeffding’s inequality aspect}
	Hoeffding’s inequality is depicted as follows:\\
	$\displaystyle Lemma 1$. Let $\displaystyle X_1,X_2,\ldots,X_n$ be independent random variables such that $\displaystyle a_i \le X_i \le b_i$ almost surely. The sum of these random variables $\displaystyle S_n=\sum X_i$ satisfies the condition that: for all $\displaystyle t>0$,
	\begin{equation}
		P(|S_n-E(S_n)|\ge t)\le2\text{exp}\bigg(-\frac{2t^2} {\sum(b_i-a_i)^2}\bigg)\notag
	\end{equation}
	The variable $\displaystyle E(S_n)$ is the expectation of $\displaystyle S_n$. The estimated sum converges to the expectation of sum by probability.
	
	From (1), since the output of each class is the sum of class-outputs of all subnetworks, Hoeffding’s inequality can be applied to our proposed growing method. 
	
	In the phase of prediction, class output of each subnetwork can be considered as a random variable. For subnetworks focus on target class $\displaystyle c_t$, we use $\displaystyle y_{c_tj}$ to represent the sum of class-output values on $\displaystyle j$-th sample of target-class $\displaystyle c_t$ and $\displaystyle y_{c_{nt}j}$ to represent the sum of class-output on $\displaystyle j$-th sample of non-target class $\displaystyle c_{nt}$. The expectation of $\displaystyle y_{c_tj}$ is represented as $\displaystyle E(y_{c_tj})$ and the expectation of $\displaystyle y_{c_{nt}j}$ of other class is represented as $\displaystyle E(y_{c_{nt}j})$. If $\displaystyle E(y_{c_tj})>E(y_{c_{nt}j})$, according to Hoeffding’s inequality, with enough subnetwork outputs $\displaystyle y_{ij}^k$ bounded by [$\displaystyle l_k, u_k$], $\displaystyle y_{c_tj}$ will converge to $\displaystyle E(y_{c_tj})$ and $\displaystyle y_{c_{nt}j}$ will converge to $\displaystyle E(y_{c_{nt}j})$ by probability as shown in (2)
	\begin{equation}
		\begin{cases}
			P(|y_{c_tj}-E(y_{c_tj})|\ge t)\le2\text{exp}\bigg(-\frac{2t^2} {\sum(u_k-l_k)^2}\bigg) \\
			P(|y_{c_{nt}j}-E(y_{c_{nt}j})|\ge t)\le2\text{exp}\bigg(-\frac{2t^2} {\sum(u_k-l_k)^2}\bigg)
		\end{cases}\tag{2}.
	\end{equation}
	Thus, the probability that $\displaystyle y_{c_{nt}j}>y_{c_tj}$ will converge to 0, there will be a clear boundary between target samples and non-target samples.
	
	The prediction phase is operated on a test set. In the selection phase, we evaluate added subnetworks on the selection set. We firstly suppose that the selection set and test set are formed by the same distribution. Therefore, the expectations $\displaystyle E(y_{c_tj})$ and $\displaystyle E(y_{c_{nt}j})$ on selection set and test set should be identical. Secondly, from Hoeffding’s inequality, with enough subnetworks, there will be a clear boundary between target samples and non-target samples. From the above supposition, we can select subnetworks with the following conditions.
	\begin{equation}
		\overline{y_{c_tj}^k}>\overline{y_{c_{nt}j}^k}\tag{3},
	\end{equation}
	\begin{equation}
		\sum_{ij}w_{ij}^ky_{ij}^k>0,\qquad w_{ij}^k=\begin{cases}
			\text{max}(\text{max}(y_{c_{nt}j})-y_{ij}, 0),&i=c_t \\
			\text{min}(\text{min}(y_{c_tj})-y_{ij}, 0),&i\neq c_t
		\end{cases}\tag{4}.
	\end{equation}
	In (3), the average class-output value on target samples should be greater than the average class-output values on non-target samples, this condition guarantees $\displaystyle y_{c_tj}>y_{c_{nt}j}$. In (4), we use a weighted sum of class-output on each sample, to make $\displaystyle \text{min}(y_{c_tj})>\text{max}(y_{c_{nt}j})$.
	
	\subsubsection{Loss aspect}
	From loss aspect, purpose of added subnetworks is to minimize loss. A new subnetwork will add value on both target samples and non-target samples. We decide which sample should be added more value and which sample should be added less value.
	
	For cross-entropy loss, we compute derivative of loss by $\displaystyle y_{ij}^k$ as show in (5) and (6).
	\begin{equation}
		-\frac{\partial l_j} {\partial y_{c_tj}^k}=1-\frac{1} {\tau_j+1}, \quad \tau_j=\frac{\sum^{N_c}_{i\neq c_t}e^{y_{ij}}} {e^{y_{c_tj}+y_{c_tj}^k}}\tag{5},
	\end{equation}
	\begin{equation}
		\frac{\partial l_j} {\partial y_{c_rj}^k}=\frac{1} {\tau_j+1}, \quad \tau_j=\frac{\sum^{N_c}_{i\neq c_r}e^{y_{ij}}} {e^{y_{c_rj}+y_{c_rj}^k}}, \quad c_r \neq c_t\tag{6}.
	\end{equation}
	In (5), if the true label is target class, for the same positive $\displaystyle y_{ij}^k$, the greater $\displaystyle \tau_j$ is, the greater loss descendent. In (6), if the true label is non-target class, the greater $\displaystyle \tau_j$ is, the smaller loss increment. Therefore, purpose of added subnetwork is to make $\displaystyle \tau_j$ as great as possible on each sample.
	
	The ideal condition of minimizing loss is that output values on all target samples is identical as $\displaystyle \overline{y_{c_tj}}$ and all output values on all non-target samples is identical as $\displaystyle \overline{y_{c_{nt}j}}$ and $\displaystyle \overline{y_{c_tj}}>\overline{y_{c_{nt}j}}$, and the variance should be minimized to 0. This condition also satisfies the condition in (4). Thus, the derivative of variance on target class and non-target class is shown in (7) and (8) respectively. 
	\begin{equation}
		-\frac{\partial Var(y_{c_tj})} {y_{c_tj}}=\frac{2} {N_{c_t}}\big(E(y_{c_tj}-y_{c_tj})\big)\tag{7},
	\end{equation}
	\begin{equation}
		-\frac{\partial Var(y_{c_{nt}j})} {y_{c_{nt}j}}=\frac{2} {N_{c_{nt}}}\big(E(y_{c_{nt}j}-y_{c_{nt}j})\big)\tag{8}.
	\end{equation}

	In conclusion, in a fixed dataset, for one added subnetwork which focus on one target class $\displaystyle c_t$. This added subnetwork should make more contribution to target samples and make contribution to lessen variance as depicted in (9).
	\begin{equation}
		\begin{cases}
			\quad \overline{y_{c_tj}^k}>\overline{y_{c_{nt}j}^k} \\
			\sum_{ij}w_{ij}^ky_{ij}^k>0,\qquad w_{ij}^k=\begin{cases}
				E(y_{c_tj}-y_{c_tj}),&i=c_t \\
				E(y_{c_{nt}j}-y_{c_{nt}j}),&i\neq c_t
			\end{cases}
		\end{cases}\tag{9}.
	\end{equation}
	
	\subsubsection{Election}
	If we use a threshold $\displaystyle thd_k$ to each added subnetwork output value as (10),
	\begin{equation}
		y_{ij}^k=\begin{cases}
			1, & y_{ij}^k>thd_k\\
			0, & y_{ij}^k\le thd_k
		\end{cases}\tag{10},
	\end{equation}
	we can leverage the form of Hoeffding’s inequality on binary distribution as shown in (11) and (12).
	\begin{equation}
		P\big(y_{c_tj}\le (p_{tj}-\varepsilon)N_n\big)\le e^{-2\varepsilon^2N_n} \qquad 0<\varepsilon<p_{tj}\tag{11},
	\end{equation}
	\begin{equation}
		P\big(y_{c_{nt}j}\ge (p_{ntj}+\varepsilon)N_n\big)\le e^{-2\varepsilon^2N_n} \qquad 0<\varepsilon<1-p_{ntj}\tag{12}.
	\end{equation}
	$\displaystyle y_{c_tj}$ is the sum of class-output on target sample $\displaystyle sp_{c_tj}$, $\displaystyle y_{c_{nt}j}$ is the sum of class-output on non-target sample $\displaystyle sp_{c_{nt}j}$. $\displaystyle p_{tj}$ is the probability that class-output on $\displaystyle sp_{c_tj}$ is 1 and $\displaystyle p_{ntj}$ is probability that class-output on $\displaystyle sp_{c_{nt}j}$ is 1.If $\displaystyle \text{min}(p_{tj})>\text{max}(p_{ntj})$ and with enough subnetwork, $\displaystyle y_{c_tj}$ will converge to $\displaystyle p_{tj}N_n$ and $\displaystyle y_{c_{nt}j}$ will converge to $\displaystyle p_{ntj}N_n$ by probability. Therefore, will form a clear boundary between target samples and non-target samples.
	
	We suppose that samples of each class are equal in selection set, we alter (9) to (13) as follows.
	\begin{equation}
		\begin{cases}
			prc_t^k>1/N_c \\
			\sum_{ij}w_{ij}^ky_{ij}^k>0, \quad w_{ij}^k=\begin{cases}
				E(y_{c_tj})-y_{c_tj}, & i=c_t\\
				E(y_{c_{nt}j})-y_{c_{nt}j}, & i\neq c_t
			\end{cases}
		\end{cases}\tag{13}.
	\end{equation}
	$\displaystyle prc_t^k$ is the precision of target samples in all the samples on which class-output is set to 1.
	
	We use this principle for election by subnetworks in trans-task case and experiment show that election mode is able to improve performance steadily without training new parameters. Detailed deduction on election is shown in \hyperlink{election}{Appendix A}.
	
	\subsection{Base network}
	We choose NAM as the base network to verify our growing method. NAM consist of multiple independent MLPs which focus on specific range of input. Output of MLP is summed as the class-output for classification. NAM is formulated as (14),
	\begin{equation}
		y_i=\sum_{k}^{n}f(x_k)[i]\tag{14},
	\end{equation}
	where $\displaystyle f(x)$ is the function of MLP, $\displaystyle x_k$ is the input of the $\displaystyle k$-th MLP. Since NAM is tree structure network, we refer each MLP as branch in the rest part of this paper. In the setting of NAM, branch represent the subnetwork in section 2.1.
	
	In this work, we firstly train an NAM as base network. Our growing method is to apply trained branches to other inputs to add branches to base network. In trans-task condition, we apply trained branches of one task to inputs of a new task to grow network in the new task.
	
	\subsection{Grow with shared parameters}
	Since we know the condition for qualified branches, we can try different branch to one input range and select qualified branches on a selection set. But randomly trying branches is not efficient in computation. In this part, we solve this problem by matching distributions.
	
	Based on theories of branch qualification, the ideal condition for matching a trained branch to one input range is that: we select one target-class $\displaystyle c_t$ on one input range and apply a trained branch to this input range, and the class-output $\displaystyle y_{ij}^k$ satisfies the condition of (9). Namely, class-output on samples of class $\displaystyle c_t$ is higher than class-output of samples of other class.
	
	\begin{figure}[h]
		\begin{center}
			\includegraphics[width=0.9\linewidth]{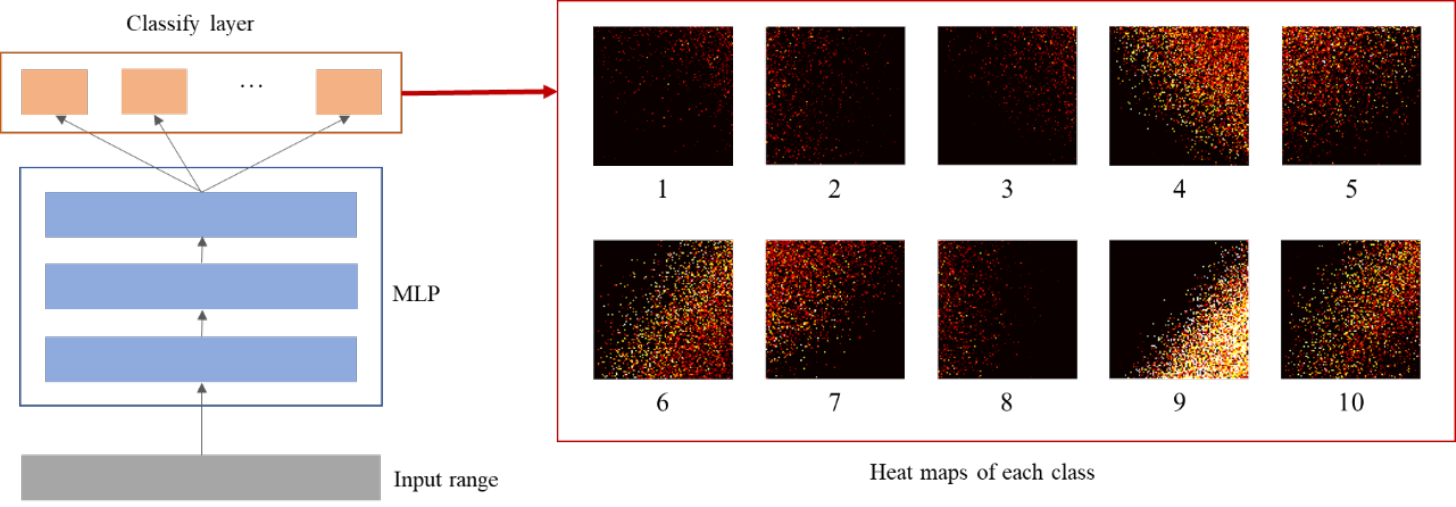}
		\end{center}
		\caption{Heat map of output distribution on each class-output, each point in heat map is a sample in the input space, color of point represents the class-output value of each class.}
	\end{figure}
	
	The problem of matching input ranges and branches is solved by matching distributions. For original trained branches, the structure and input-output distribution of each class are shown in Figure 3, we randomly select two dimensions from the input space, and each point in the input space is an input sample. In the distribution heat map, input sample is randomly generated from uniform distribution. For each class, some part of input space obtains higher output value and other part obtains lower output value, namely, each class focuses on specific part of input space. For input range to be matched, we select one class as target class and other class as non-target class, the distribution is shown in the left scatter map of Figure 4. Our purpose for matching is to find a candidate branch, target samples matches the branch’s high-value part as shown in Figure 4, and make class-output on target samples higher than class-output on other samples. The matched branch only focuses on the target class, namely, the red point in Figure 4.
	
	\begin{figure}[h]
		\begin{center}
			\includegraphics[width=0.8\linewidth]{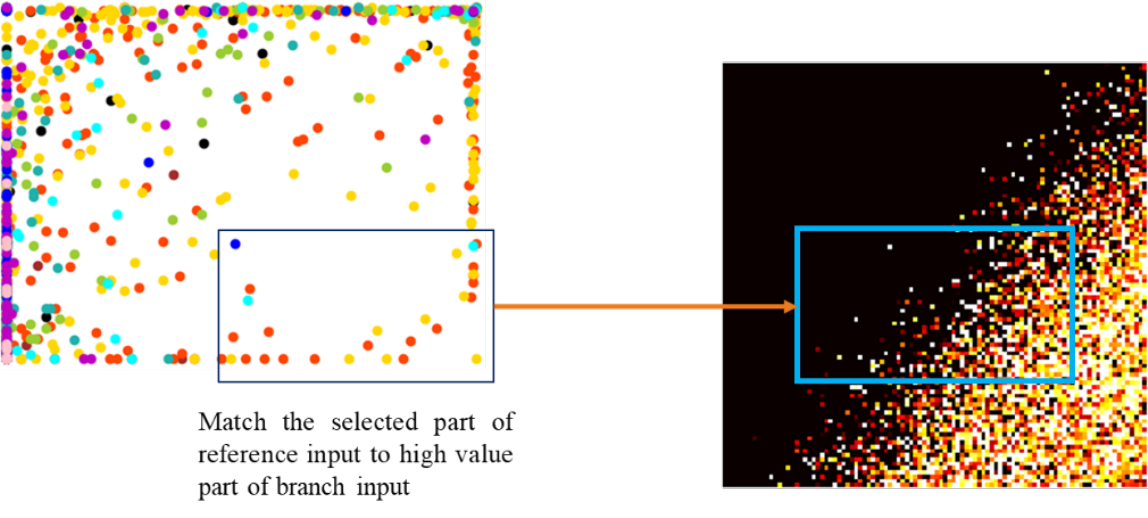}
		\end{center}
		\caption{Matching red point to high value part of heat map to make class-output on red point higher than class-output on other values.}
	\end{figure}
	
	To make a clear definition, input range to be matched is referred as reference input, sample of reference input is referred as reference sample. The class to be focused on reference input is referred as target class $\displaystyle c_t$, other class is referred as non-target class. Branch to be matched is referred as candidate branch, class which is selected for class-output is referred as branch-class $\displaystyle c_b$.
	
	With the theoretical base of branch qualification and the purpose of matching branches, the overall process is summarized in Figure 5. We grow neural network in an iterative fashion. In the clustering step, we transfer the distribution of high value part in Figure 2 to a set of clusters. In the matching step, we match target sample to clusters of high value part and make target samples obtain higher class-output values. In the branch selection step, we select qualified branches by (9) and (13). in the steps of adding network, we show how to add branches to current network for further tuning or election.
	
	\begin{figure}[h]
		\begin{center}
			\includegraphics[width=0.9\linewidth]{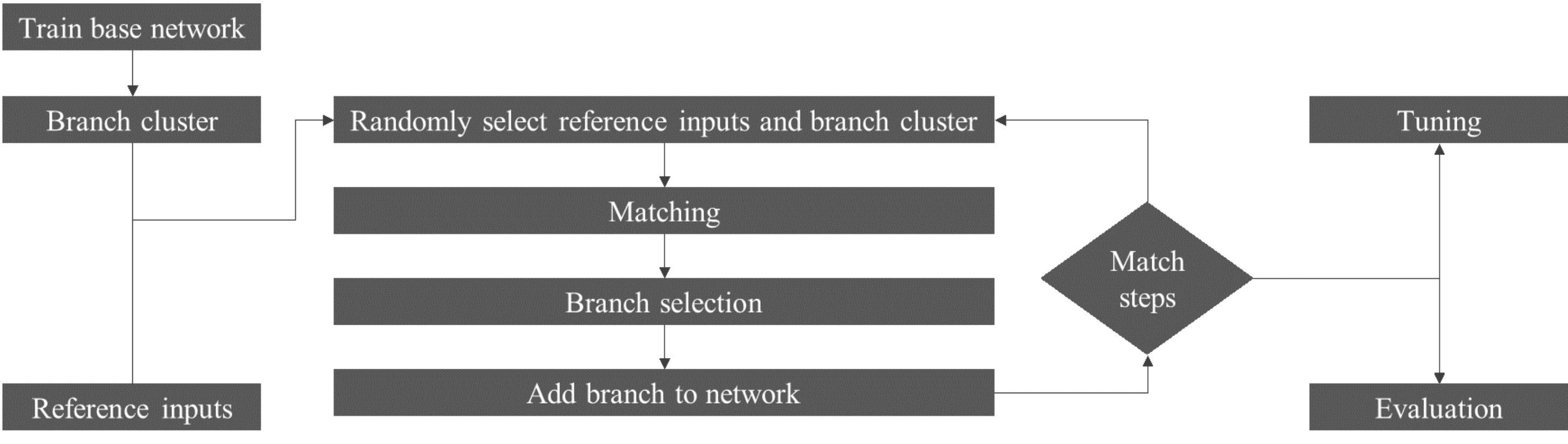}
		\end{center}
		\caption{Overall process of our growing method.}
	\end{figure}
	
	\subsubsection{Clustering}
	We consider the high value part in Figure 3 as the combination of multiple clusters, and clustering method convert high-value distribution to combination of clusters.
	
	We do cluster on each class in an iterative way. First, we use \emph{multi-variate Gaussian kernel} to find the densest point in the branch input space. Then we use nearest neighbor method to search all the samples of this cluster and form a cluster. Finally, we remove the selected samples and repeat the former steps to find new clusters.
	
	Each cluster is a pair contains class $\displaystyle c_{b_i}$ cluster center $\displaystyle sp_c$ max class-output value $\displaystyle y_{c_{bij}max}$ as $\displaystyle \{c_{b_i}, sp_c, y_{c_{bij}max}\}$. We depict a cluster as a center with its maximum class-output. Detailed implementation of clustering is shown in \hyperlink{clustering}{Appendix B}.
	
	\subsubsection{Matching}
	With branch clusters, we can match reference samples to branch clusters. We assume if one reference input point is closer to the center of one cluster, the class-output will be greater.
	
	Based on this supposition, we can evaluate matching by distance. For branch class $\displaystyle c_{b_i}$ with multiple cluster $\displaystyle Cl_{c_{bi}j}$ and reference class $\displaystyle c_{r_i}$ with $\displaystyle N$ reference samples $\displaystyle sp_{r_ij}$, partial average distance is defined in (15), 
	\begin{equation}
		d_{b_ir_i}=\frac{1} {n}\sum_{j}w_{nst}\sqrt{(sp_{c_{ri}j}-Cl_{nst})^2}, \quad w_{nst}=\frac{e^{y_{nstmax}}} {\sum_{j}e^{y_{c_{bij}max}}}, \quad n\le N\tag{15}.
	\end{equation}
	$\displaystyle Cl_{nst}$ is the nearest center from $\displaystyle sp_{r_ij}$, $\displaystyle y_{nstmax}$ is the max class-output value of the nearest cluster, $\displaystyle n\le N$ means if there are $\displaystyle n$ sample satisfies the matching condition that $\displaystyle d_{b_ir_i}$ is the smallest distance among all reference classes, we set a boundary in input space and drop samples out of boundary. We iteratively evaluate combinations of all $\displaystyle c_{b_i}$ and $\displaystyle c_{r_i}$, and find the best matched reference class for each branch class.
	
	After finding the best-matched candidate branch and corresponding branch class $\displaystyle c_{b_i}$ on one reference class $\displaystyle c_{r_i}$, we add this branch to candidate branches for branch selection step, and the reference class $\displaystyle c_{r_i}$ is marked as target-class $\displaystyle c_t$ of new branch. Only class output of $\displaystyle c_{b_i}$-th class will be used as class output of the matched branch and class-output of other branch class is set to 0.
	
	We also developed parameter transfer method to make matching easier as show in \hyperlink{parameter transfer}{Appendix C}.
	
	\subsubsection{Branch selection}
	Matched branches will be evaluated in a selection dataset, which is a subset randomly sampled from training dataset. The qualified branches satisfy the conditions in (9) and (13) in branch qualification part.
	
	For the qualification condition, we use (13) to check each added branch. We also use a threshold value to select class-output values greater than this threshold. Class-output values lower than threshold are set to 0. To make the contributions of target class greater than non-target class, there will be a non-negative bias added to all the above-threshold samples in the tuning step.
	
	Qualified added branches will be used on further tuning or election.
	
	\subsubsection{Tuning}
	After branch qualification, we add a class-mask which is formulated in (16) to each added branch. 
	\begin{equation}
		y_{ij}^{k\prime}=\text{ReLU}(a)\bigg(\text{ReLU}\Big(\frac{y_{ij}^k-thd} {v_{span}}\Big)+s(y_{ij}^k)\text{ReLU}(b)\bigg), \quad s(y_{ij}^k)=\begin{cases}
			1, & y_{ij}^k>thd \\
			0, & y_{ij}^k\le thd
		\end{cases}\tag{16},
	\end{equation}
	the variable $\displaystyle v_{span}$ is the value span $\displaystyle \text{max}(y_{ij}^k)-thd$ computed in the selection set and applied in tuning and testing, $\displaystyle a$ and $\displaystyle b$ are trainable variable as scale and bias, scale and bias are set to positive using ReLU. This setting is to make each branch make more contribution to target samples than non-target samples. As shown in (16), only scale and bias will be trained in the tuning mode, and once class mask is trained, parameters is fixed and won’t be trained later, this makes tuning efficient.
	
	\subsubsection{Election}
	In the election mode, we compute class score as (17). 
	\begin{equation}
		score_{ij}=\sum_{k}\frac{y_{ij}^k-E(y_{ij}^k)} {\sigma(y_{ij}^k)}\tag{17},
	\end{equation}
	$\displaystyle E(y_{ij}^k)$ and $\displaystyle \sigma(y_{ij}^k)$ is the mean and std of $\displaystyle y_{ij}^k$. These two values are computed on training set and applied in test set. The prediction of class is the class which has the maximum score among all class scores.
	
	\section{Experiment}
	We test our growing method on two tasks: 1) adding new branches to base network to improve the performance of original network, we show our growing method can gradually improve performance with higher parameter efficiency. 2) We also use branches in base network to do election on another task, and we show our growing method can gradually improve performance in trans-task case. Our growing method can transfer network without training.
	
	\subsection{Base network}
	We train base network on cifar10 dataset. Cifar10 dataset which contains 50000 training samples and 10000 testing samples is used for image classification task. We normalize all samples to the range of [-0.5, 0.5] and no data augmentation is applied. Each image in cifar10 is of [3, 32, 32]. We set each branch of base network percept the input range of $\displaystyle [6i:6i+3;6j:6j+3], i,j\epsilon N$ of each channel, each branch contains 4 layers and each layer is a fully connected layer with dimension 9. Base network originally contains 75 branches. We train the base network by gradient descend method and use Adam optimizer for optimization.
	
	\subsection{Network growth}
	After training base network, we apply our growing method to the base network. We add branches to all the input ranges $\displaystyle [i:i+3;j:j+3], i,j\epsilon N$. Figure 6 depicts the accuracy and loss on test set with the increment of added branches. We can see a steady increment on accuracy from 0.468 to 0.5016 and a steady decrement on loss from 1.4684 to 1.4379. To compare the parameter efficiency, we also trained a full-perception NAM which percept all the input ranges and all channels of input image as the comparison network of NAM on cifar10 task, performance and parameters of full perception network is shown in Table 1. The grown network added totally 1456 branches, each added branch contains 2 new parameters of class-mask. Therefore, training parameters of grown network is 36662, which is much less than parameters of full perception network, and the performance of grown network is comparable with full perception network.
	
	\begin{table}[t]
		\caption{performance and parameters on cifar10}
		\label{table-1}
		\begin{center}
			\begin{tabular}{llll}
				\multicolumn{1}{c}{\bf Network}  &\multicolumn{1}{c}{\bf Accuracy}  &\multicolumn{1}{c}{\bf Loss}  &\multicolumn{1}{c}{\bf parameters}
				\\ \hline \\
				Base network         &0.468    &1.4684   &33750\\  
				Full perception network       &0.5022  &1.3873  &135000 \\
			\end{tabular}
		\end{center}
	\end{table}
	
	From Figure 6, we can see a steady improvement on performance, which verified the effectiveness of our growing method. By comparing the performance of grown NAM with the full-perception NAM, the performance of grown network has shown a higher parameter efficiency.
	
	\begin{figure}[h]
		\begin{center}
			\includegraphics[width=0.7\linewidth]{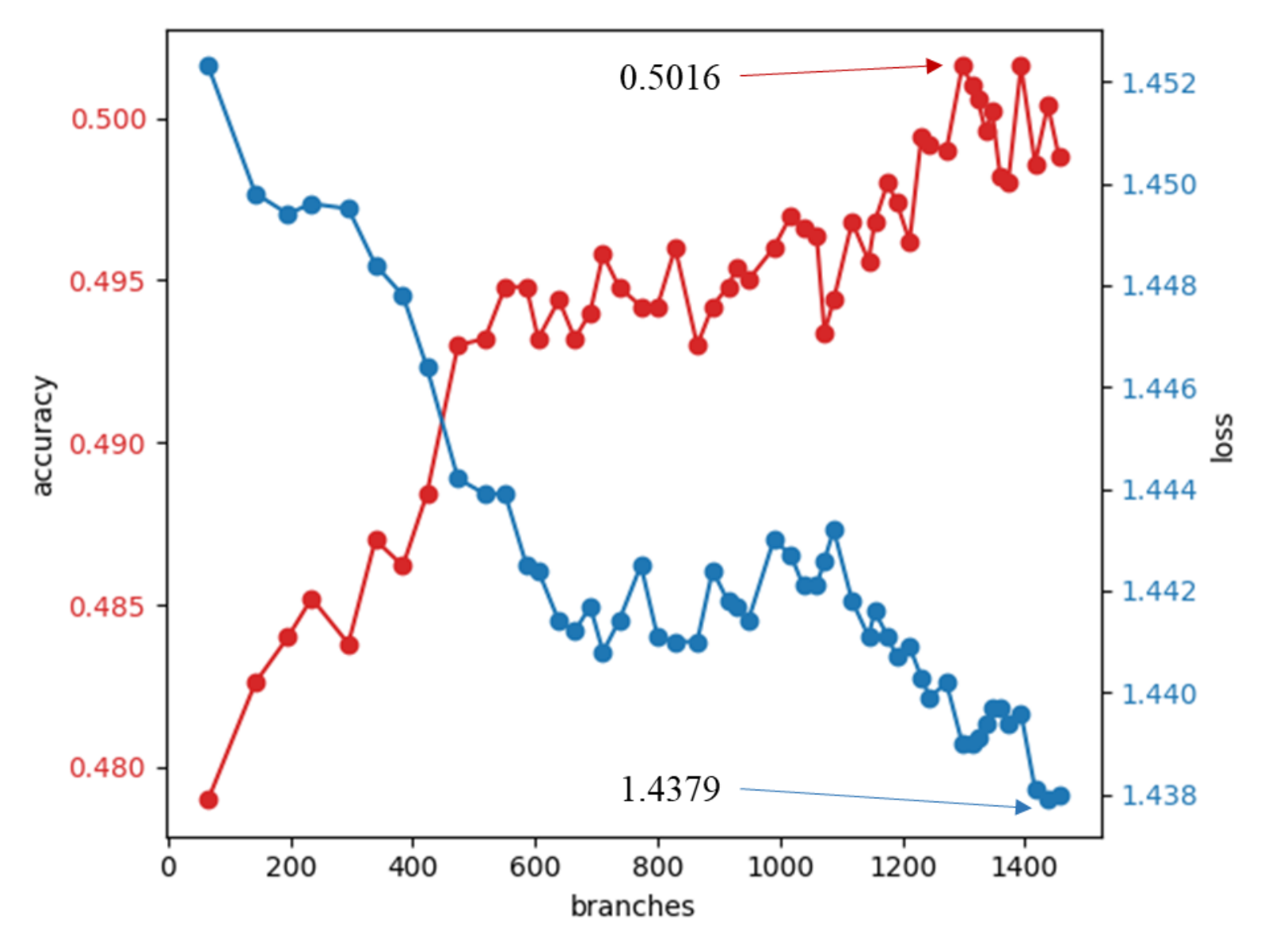}
		\end{center}
		\caption{Loss and accuracy of grown network as new branches added.}
	\end{figure}
	
	\subsection{Transfer task learning}
	After training the base network on cifar10 dataset, we apply branches of base network to MNIST dataset. MNIST is a dataset which contains 50000 training samples and 10000 test samples and used for written number recognition. We normalize each sample to range [-0.5, -.05] and there is no data augmentation.
	
	\begin{figure}[h]
		\begin{center}
			\includegraphics[width=0.7\linewidth]{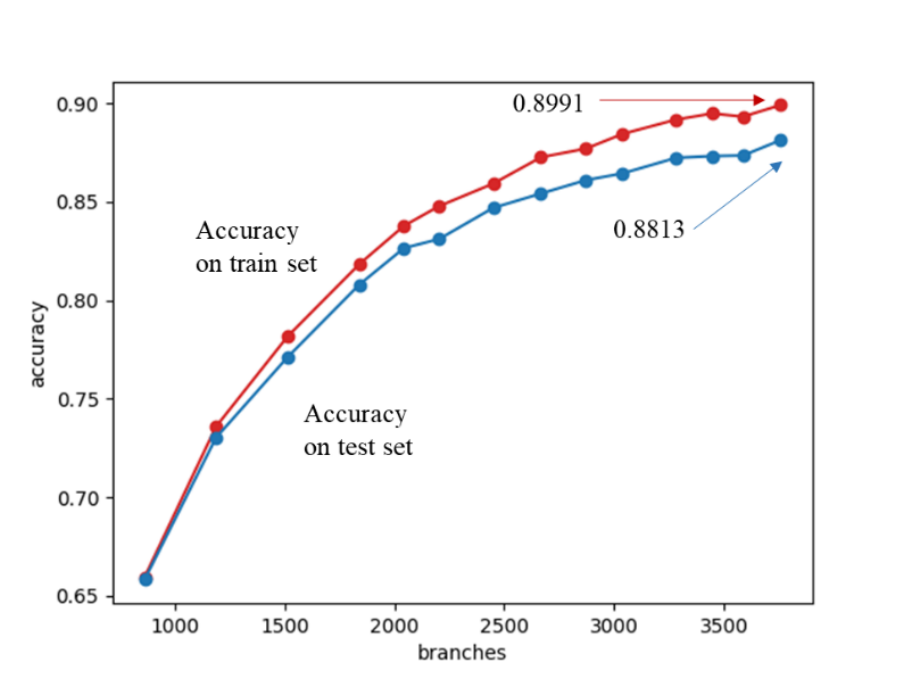}
		\end{center}
		\caption{Accuracy of grown network on train set and test set as new branches added.}
	\end{figure}
	
	There is initially no branch on MNIST task, and we gradually apply the originally trained 75 branches of base network to this task, and use election mode for prediction. Figure 7 depicts the accuracy on test set with the increment of branches. From Figure 7, there is a steady increment on the performance and achieve 0.8991 on train set and 0.8813 on test set. This performance is realized by applying the original 75 branches to different input ranges of MNIST task.
	
	The experiment indicates that: by applying branches in different input range, branches trained on one task can be applied to another task without training new parameters, and changing function of one neural network can be realized by perturbating the order of subnetworks.
	
	To compare the representability of NAM on MNIST task, we train a full perception NAM, the performance is shown in Table 2. There is a gap of 0.08 on accuracy, which means our current trans-task growing is not better than training directly.
	
	\begin{table}[t]
		\caption{performance and parameters on MNIST}
		\label{table-2}
		\begin{center}
			\begin{tabular}{llll}
				\multicolumn{1}{c}{\bf Network}  &\multicolumn{1}{c}{\bf Accuracy}  &\multicolumn{1}{c}{\bf Loss}  &\multicolumn{1}{c}{\bf parameters}
				\\ \hline \\
				Full perception network       &0.9613  &0.1362  &36450 \\
			\end{tabular}
		\end{center}
	\end{table}
	
	\section{Related works}
	There have been several works focus on growing neural networks to gradually improve performance on a given task. (\cite{schwenk2000boosting}; \cite{bengio2006convex}; \cite{bach2017breaking}) progressively adding new parameters to existing neural network by the indication of gradient while keeping previous parameters fixed. However, this set of works need to solve complex non-convex optimization problems. Our growing method also add new branches with previous parameters fixed, but leveraging statistic base to avoid optimization problems. 
	
	Another set of growing methods grow networks by splitting neurons, (\cite{wynne1991node}; \cite{chen2015net2net}) grow neural network by heuristic or random strategies, (\cite{liu2019splitting}) propose the Splitting Steepest Descendent which provide a new method to selectively split neurons while guarantee the improvement of performance, but the growing operation is limited to splitting neurons. (\cite{wu2021firefly}) further extent the growing operations to both deepen and widen neural networks and optimize the grown network in a more efficient way. These splitting methods is efficient on growing neural network by the indication of loss. However, these methods need to introduce new training parameters to existing neural network.
	
	Neural architecture search (NAS) methods also including adding new parameters to existing neural network. (\cite{hu2019efficient}) proposed an efficient way of growing neural networks by adding shortcuts as gradient boosting. In our work, we use both gradient indication and statistic indication.
	
	\section{Conclusion}
	In this work, we propose a method for growing network with shared parameter and the theoretical base of our growing method. Our growing method provide a possible way to dynamically improve the performance of a given network with few added parameters, and a new approach for transfer learning without tuning parameter but change the combination of different parameters. For future work, we will make growth on add new layers, do matching based on attention models, make growth more efficient and apply our growing method to more tasks. Source code of our growing method can be found in https://github.com/Rain-axt/Growing-neural-networks.

	\appendix
	\section{Appendix}
	\hypertarget{election}{detailed deduction of election:}\\
	
	Let  $\displaystyle X_1,X_2,\ldots,X_n$ be independent random variables such that $\displaystyle a_i \le X_i \le b_i$ almost surely. The average value of these random variables $\displaystyle \overline{X}=\frac{1} {n}\sum_{i}X_i$ satisfies the condition that: for all $\displaystyle t>0$:\\
	\begin{equation}
		P(\overline{X}-E(\overline{X})\ge t)\le \text{exp}\bigg(-\frac{2n^2t^2} {\sum_{i}(b_i-a_i)^2}\bigg)\notag,
	\end{equation}
	if each random variable is limited within {0, 1} and satisfy distribution in (A.1),\\
	\begin{equation}
		P(X_i=1)=p_1\tag{A.1},
	\end{equation}
	The average form of Hoeffding’s inequality can be transferred to (A.2)
	\begin{equation}
		P(\overline{X}-p_1n\ge t)\le \text{exp}(-2nt^2)\tag{A.2}.
	\end{equation}

	If each added subnetwork output value is limited within {0, 1}, given $\displaystyle y_{c_tj}$ is the sum of class-output of target sample $\displaystyle sp_{c_tj}$, $\displaystyle y_{c_{nt}j}$ is the sum of class-output of non-target sample $\displaystyle sp_{c_{nt}j}$, $\displaystyle p_{tj}$ is the probability that class-output on $\displaystyle sp_{c_tj}$ is 1 and $\displaystyle p_{ntj}$ is probability that class-output on $\displaystyle sp_{c_{nt}j}$ is set to 1. According to (A.2), we have (11) and (12).\\
	\begin{equation}
		P\big(y_{c_tj}\le (p_{tj}-\varepsilon)N_n\big)\le e^{-2\varepsilon^2N_n} \qquad 0<\varepsilon<p_{tj}\tag{11}
	\end{equation}
	\begin{equation}
		P\big(y_{c_{nt}j}\ge (p_{ntj}+\varepsilon)N_n\big)\le e^{-2\varepsilon^2N_n} \qquad 0<\varepsilon<1-p_{ntj}\tag{12}
	\end{equation}

	Consider there is totally $\displaystyle N_k$ samples on which class-output is set to 1 from the $\displaystyle k$-th added subnetwork, the precision of target class $\displaystyle prc_t^k$ and on non-target class $\displaystyle prc_{nt}^k$ is defined as (A.3),\\
	\begin{equation}
		prc_t^k=\frac{N_t^k} {N_k}, \quad prc_{nt}^k=\frac{N_{nt}^k} {N_k}\tag{A.3}.
	\end{equation}
	For all the subnetworks, the precision is defined as (A.4),\\
	\begin{equation}
		prc_t=\frac{N_t} {N}, \quad prc_{nt}=\frac{N_{nt}} {N}\tag{A.4},
	\end{equation}
	where $\displaystyle N$ $\displaystyle N_t$ and $\displaystyle N_{nt}$ is the sum of $\displaystyle N_k$ $\displaystyle N_t^k$ and $\displaystyle N_{nt}^k$ respectively.
	
	From (8) and (9), variance of class-output on target samples is minimized to 0 and class-output non target samples is also minimized to 0. In the ideal condition, precision of all non-target class will be identical. Therefore, we have the relation between $\displaystyle prc_t$ and $\displaystyle prc_{nt}$ as (A.5).\\
	\begin{equation}
		prc_{nt}=\frac{1-prc_t} {N_c-1}\tag{A.5}.
	\end{equation}
	To make a clear boundary between target samples and non-target samples, precisions should satisfy $\displaystyle prc_t>prc_{nt}$, substitute (A.5) to this condition, we have condition of $\displaystyle prc_t$ as show in (A.6).
	\begin{equation}
		prc_t>\frac{1} {N_c}\tag{A.6}.
	\end{equation}
	
	We expand this condition to each subnetwork precision and keep minimizing variance of class-output on target samples and non-target samples. The final condition is depicted as (13).\\
	\begin{equation}
		\begin{cases}
			prc_t^k>1/N_c \\
			\sum_{ij}w_{ij}^ky_{ij}^k>0, \quad w_{ij}^k=\begin{cases}
				E(y_{c_tj})-y_{c_tj}, & i=c_t\\
				E(y_{c_{nt}j})-y_{c_{nt}j}, & i\neq c_t
			\end{cases}
		\end{cases}\tag{13}.
	\end{equation}

	\section{Appendix}
	\hypertarget{clustering}{detailed implementation of clustering:}\\
	
	We firstly generate samples which are randomly drawn from uniform distribution for candidate branches, we compute class-output on each class and save the pairs \{sample, class, class-output\} as \{$\displaystyle sp_j$, $\displaystyle c_{b_i}$, $\displaystyle y_{b_ij}$\}, we refer these pairs as branch-pair. For each branch-class $\displaystyle c_{b_i}$, only pairs with top 20\% class-output value $\displaystyle y_{b_ij}$ are reserved and used for clustering. Reference samples is selected from train-set, the labels of reference samples are the true labels $\displaystyle c_{tr}$ from dataset, reference pairs \{sample, true-class\} are saved as \{$\displaystyle sp_j$, $\displaystyle c_{tr}$\}.
	
	\begin{table}[t]
		\caption{Pseudo code of clustering}
		\label{table-b.1}
		\begin{tabular}{l}
			\textbf{Algorithm 1:} Clustering
			\\ \hline \\
			\textbf{input}: set of samples of class $\displaystyle c_{b_i}$ $\displaystyle S=$\{$\displaystyle sp_j$, $\displaystyle c_{b_i}$, $\displaystyle y_{b_ij}$\}; neighbor distance $\displaystyle d_{nb}$; minimum shift \\
			\qquad distance $\displaystyle d_{min}$;empty cluster set $\displaystyle S_{cl}$\\
			\textbf{While} $\displaystyle |S|>0$ \textbf{do} \\
			\qquad Random select start point \bm{$\displaystyle sp_{cr}$} \\
			\qquad \textbf{While} $\displaystyle shift\_distance > d_{min}$ \textbf{do} \\
			\qquad \qquad Compute shifted point \bm{$\displaystyle sp_{sft}$} by (B.2) \\
			\qquad \qquad $\displaystyle \text{dist}($\bm{$\displaystyle sp_{cr}, $}\bm{$\displaystyle sp_{sft}$}$\displaystyle ) \rightarrow shift\_distance$ \\
			\qquad \qquad bm{$\displaystyle sp_{sft}$} $\displaystyle \rightarrow$ \bm{$\displaystyle sp_{cr}$} \\
			\qquad \textbf{End} \\
			\qquad Select neighbor samples \{\bm{$\displaystyle sp_{nb}$} | $\displaystyle \text{dist}($\bm{$\displaystyle sp_{cr}, $}\bm{$\displaystyle sp_{sft}$}$\displaystyle )<d_{nb}$\} as cluster \\
			\qquad Add cluster to $\displaystyle S_{cl}$ \\
			\qquad Remove neighbor samples from $\displaystyle S$ \\
			\textbf{End} \\
			\hline \\
		\end{tabular}
	\end{table}
	
	In this work, we use mean shift cluster (MSC) as our clustering method. The pseudo code of clustering is shown in Table B.1. Mean shift cluster is an iterative method to shift one point to denser point and gradually approach to the densest point. We use \emph{multi-variate Gaussian kernel} to compute density weight for each sample as depicted in (B.1).\\
	\begin{equation}
		G(\boldsymbol{sp_1},\boldsymbol{sp_2})=\frac{1} {2\pi^{\frac{1} {n}}|cov|^{-1}}\text{exp}\big((\boldsymbol{sp_1}-\boldsymbol{sp_2})^Tcov^{-1}(\boldsymbol{sp_1}-\boldsymbol{sp_2})\big)\tag{B.1}.
	\end{equation}
	The shifted point is the weighted sum of all current samples as shown in (B.2). \\
	\begin{equation}
		\boldsymbol{sp_{sft}}=\sum_{j}\frac{G(\boldsymbol{sp_{cr}},\boldsymbol{sp_j})} {Z}\boldsymbol{sp_j}, \quad Z=\sum_{j}G(\boldsymbol{sp_{cr}},\boldsymbol{sp_j})\tag{B.2}.
	\end{equation}

	We stop shifting point if the shifted point is close enough to current point, and current point is the densest point. After finding the densest point, we select neighbor point within a distance and form a cluster of samples.
	
	Each class of branch pairs are clustered separately. To make clustering suitable for different distribution, we normalize branch samples to 0-mean and 1-std before clustering and do denormalization after clustering.
	
	For each cluster, the max class-output value and corresponding sample are marked as center of this cluster, which means higher output value is near this center. We represent one cluster $\displaystyle Cl_{c_{bi}j}$ as \{$\displaystyle c_{b_i}$, $\displaystyle sp_c$, $\displaystyle y_{c_{bij}max}$\}, each cluster is a pair contains class $\displaystyle Cl_{c_{bi}j}$ cluster center $\displaystyle sp_c$, max class-output value $\displaystyle y_{c_{bij}max}$. We depict a cluster as a center with its maximum class-output.

	\section{Appendix}
	\hypertarget{parameter transfer}{detailed implementation of parameter transfer:}\\
	
	To make matching easier, we normalize both branch clusters and reference samples. Normalization will make patterns of branch clusters and reference samples closer.
	
	For branch clusters $\displaystyle Cl_i$ of one branch class $\displaystyle c_{b_i}$, there is multiple samples within each cluster and we use $\displaystyle set(x_{bi})$ to represent the sample set, and each cluster is pair of {class, center, max class-output} \{$\displaystyle c_{b_i}$, $\displaystyle sp_c$, $\displaystyle y_{c_{bij}max}$\}. We normalize clusters of $\displaystyle c_{b_i}$ as (C.1),\\
	\begin{equation}
		\begin{cases}
			sp_{ci}^\prime=\frac{sp_{ci}-\overline{x_{bi}}} {\text{max}(x_{bi})-\text{min}(x_{bi})}[d_1,d_2,\ldots,d_m]\\
			\overline{x_{bi}[d_{k1}]}\le \overline{x_{bi}[d_{k2}]}, \qquad k_1<k_2
		\end{cases}\tag{C.1}.
	\end{equation}
	The operation of [] is the dimension perturbation which reorder dimensions of each cluster-center. we sort each dimension to make average values increase by dimensions. $\displaystyle sp_{ci}^\prime$ is the normed center.
	
	For reference samples $\displaystyle sp_{r_ij}$ of one reference class $\displaystyle c_{ri}$, we apply the same normalization method to normalize each sample. It worth noting that, the dimension order of cluster and reference samples are not identical.
	
	After matching candidate branch and reference input, we applied a parameter transfer step to counter-act normalization. If one point in the reference space matches the point of branch-input space as shown in (C.2),\\
	\begin{equation}
		\begin{cases}
			\frac{x_{ri}[i_r]-\overline{x_{ri}[i_r]}} {\sigma(x_{ri}[i_r])}=\frac{x_{bi}[i_b]-\overline{x_{bi}[i_b]}} {\sigma(x_{bi}[i_b])} \\ \sigma(x_{ri}[i_r])=\text{max}(x_{ri})[i_r]-\text{min}(x_{ri})[i_r] \\
			\sigma(x_{bi}[i_b])=\text{max}(x_{bi})[i_b]-\text{min}(x_{bi})[i_b]
		\end{cases}\tag{C.2},
	\end{equation}
	and the computation of the first layer of matched branch is shown in (C.3),\\
	\begin{equation}
		y_{d_i}=\sum_{i_b}w_{d_ii_b}x_{bi}[i_b]+b_{d_i}\tag{C.3},
	\end{equation}
	the purpose of parameter transfer is to find a transferred parameter to satisfy the condition in (C.4),\\
	\begin{equation}
		\sum_{i_r}w_{d_ii_r}^\prime x_{ri}[i_r]+b_{d_i}^\prime=y_{d_i}\tag{C.4},
	\end{equation}
	namely, the output of branch-input sample should be identical to output of reference sample. 
	
	We can make a transformation from $\displaystyle x_{ri}[i_r]$ to $\displaystyle x_{bi}[i_b]$ as  as (C.5). 
	\begin{equation}
		x_{bi}[i_b]=\frac{\sigma(x_{bi}[i_b])(x_{ri}[i_r]-\overline{x_{ri}[i_r]})} {\sigma(x_{ri}[i_r])}+\overline{x_{bi}[i_b]}\tag{C.5}
	\end{equation}
	By substituting (C.5) into (C.3), we have (C.6).
	\begin{equation}
		y_{di}=\sum_{i_b}w_{d_ii_b}\frac{\sigma(x_{bi}[i_b])x_{ri}[i_r]} {\sigma(x_{ri}[i_r])}-\sum_{i_b}w_{d_ii_b}\frac{\sigma(x_{bi}[i_b])\overline{x_{ri}[i_r]}} {\sigma(x_{ri}[i_r])}+\sum_{i_b}w_{d_ii_b}\overline{x_{bi}[i_b]}+b_{d_i}\tag{C.6}.
	\end{equation}
	Form (C.6), we can obtain the transfer weight and bias as (C.7).
	\begin{equation}
		\begin{cases}
			\qquad w_{d_ii_r}^\prime=w_{d_ii_b}\frac{\sigma(x_{bi}[i_b])} {\sigma(x_{ri}[i_r])}\\
			b_{d_i}^\prime=b_{d_i}-\sum_{i_b}w_{d_ii_b}\frac{\sigma(x_{bi}[i_b])\overline{x_{ri}[i_r]}} {\sigma(x_{ri}[i_r])}+\sum_{i_b}w_{d_ii_b}\overline{x_{bi}[i_b]}
		\end{cases}\tag{C.7}.
	\end{equation}

	This step of parameter transfer is not necessary for matching but will greatly improve the efficiency of matching. In practice, we apply parameter transfer to accelerate matching.
	
\end{document}